\documentclass[letterpaper, 10 pt, journal, twoside]{ieeetran}  

\usepackage{amsmath}
\usepackage{pifont}
\usepackage{amssymb}
\usepackage{bbm}
\usepackage{graphicx}
\usepackage{glossaries}
\usepackage{xcolor}
\usepackage{dsfont}
\usepackage[normalem]{ulem}

\definecolor{red}{rgb}{1., 0., 0.}
\definecolor{fernandoblue}{RGB}{0,0,0}

 \definecolor{finalcolor}{RGB}{0, 0, 0}
\newcommand{\cmark}{\ding{51}}%
\newcommand{\xmark}{\ding{55}}%
\newcommand{\fernando}[1]{{\color{fernandoblue} #1}}
\newcommand{\claude}[1]{{\color{fernandoblue} #1}}
\newcommand{\final}[1]{{\color{finalcolor} #1}}
\newcommand{\maxspeed}{$13\,\text{m/s}$ }
\newcommand{\successrate}{$81\%$ }

\newcommand{\inferencetime}{$20\,\text{ms}$}

\newcommand{\papertitle}{EV-Catcher: High-Speed Object Catching Using Low-latency Event-based Neural Networks}

\title{\LARGE \bf\papertitle}

\author{Ziyun Wang$^{*1}$, Fernando Cladera Ojeda$^{*1}$, Anthony Bisulco$^{12}$, Daewon Lee$^{2}$, \\Camillo J. Taylor$^{1}$, Kostas Daniilidis$^{1}$, M. Ani Hsieh$^{1}$, Daniel D. Lee$^{2}$, and Volkan Isler$^{2}$
\thanks{Manuscript received: February, 24, 2022; Revised May, 21, 2022; Accepted June, 15, 2022;  Date of current version \today. This paper was recommended for publication by Editor Eric Marchand upon evaluation of the Associate Editor and Reviewers' comments. *\emph{Ziyun Wang and Fernando Cladera Ojeda contributed equally to this work. (Corresponding author: Ziyun Wang)}}%
\thanks{$^{1}$Z. Wang, F. Cladera Ojeda, A. Bisulco, C. J. Taylor, K. Daniilidis, and M. A. Hsieh are with GRASP Laboratory, University of Pennsylvania {\tt\small\{ziyunw, fclad, abisulco, cjtaylor, kostas, mya\}@seas.upenn.edu}.}%
\thanks{$^{2}$A. Bisulco, D. Lee, D. D. Lee, and V. Isler are with  the Samsung AI Center NY, 837 Washington Street, New York, New York 10014. {\tt \small \{daewon.l, daniel.d.lee, ibrahim.i\}@samsung.com}.}
\thanks{Digital Object Identifier (DOI): see top of this page.}
}

\begin{document}

\maketitle

\newacronym{dvs}{DVS}{Dynamic Vision Sensor}
\newacronym{cis}{CIS}{Conventional CMOS-based image sensors}
\newacronym{uav}{UAV}{Unmanned Aerial Vehicle}
\newacronym{led}{LED}{Light-Emitting Diode}
\newacronym{som}{SOM}{System-on-module}
\newacronym{ttc}{TTC}{time-to-collision}
\newacronym{behi}{BEHI}{Binary Event History Image}


\begin{abstract}
Event-based sensors have recently drawn increasing interest in robotic perception due to their lower latency, higher dynamic range, and lower bandwidth requirements compared to standard CMOS-based imagers. These properties make them ideal tools for real-time perception tasks in highly dynamic environments. 
In this work, we demonstrate an application where event cameras excel: accurately estimating the impact location of fast-moving objects. 
We introduce a lightweight event representation called {\it Binary Event History Image} (BEHI) to encode event data at low latency, as well as a learning-based approach that allows real-time inference of a confidence-enabled control signal to the robot. 
To validate our approach, we present an experimental catching system in which we catch fast-flying ping-pong balls.
We show that the system is capable of achieving a success rate of $81\%$ in catching balls targeted at different locations, with a velocity of up to \maxspeed 
even on compute-constrained embedded platforms such as the Nvidia Jetson NX. 
\end{abstract}

\IEEEpeerreviewmaketitle

\section{INTRODUCTION}

\begin{figure}[!t]
    \centering
    \includegraphics[width=\linewidth]{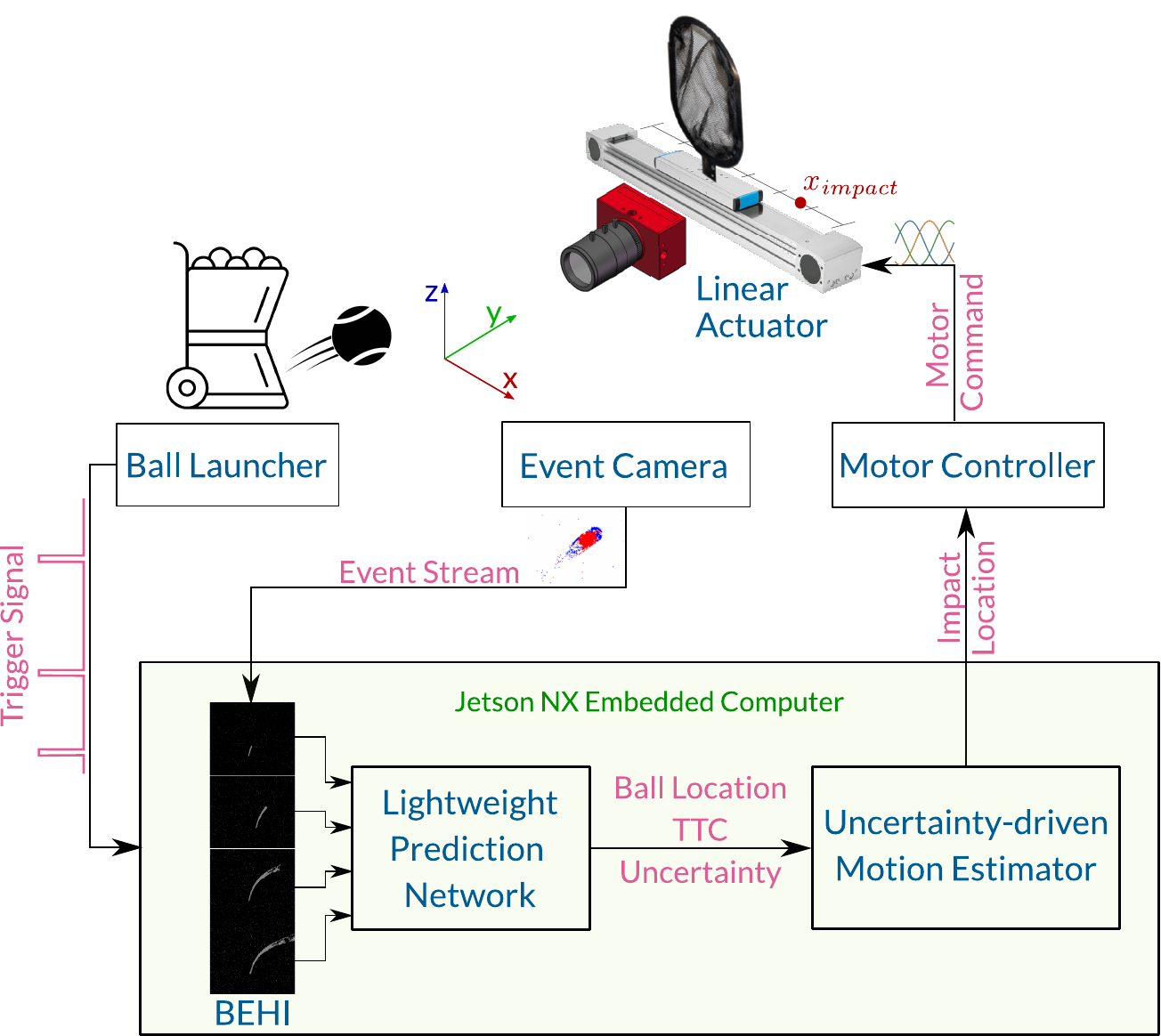}
    \caption{Experimental setup used to validate our approach: an event-based camera observes events from the incoming ball.  These events are packaged into \gls{behi} images and sent to the lightweight prediction network. The prediction network produces trajectory estimations with uncertainty. The robust motion estimator takes the network output and predicts the impact location, which is passed to the motor controller 
    for commanding a linear actuator to catch the ball. }
    \label{fig:setup}
    \vspace{-.5cm}
\end{figure}

Biological systems are able to estimate and catch objects moving at very fast speeds. For example, professional table tennis players can hit table tennis balls at speeds greater than 25 $\frac{m}{s}$, and Major League Baseball catchers can catch \fernando{fast balls} flying toward them at a speed of 40 $\frac{m}{s}$~\fernando{\cite{nathan2012analysis}}. 

On the other hand, state-of-the-art robots can only catch objects at lower speeds using vision-based sensors, as shown in Tab.~\ref{tab:catching_methods}~\cite{deguchi2008goal, lin2019ball, sato2020high, zhang2011tracking, 7018920} . 
This performance difference can be mainly explained by the limitation of robot perception using frame-based cameras: with high-speed motion, traditional cameras can only receive a few frames within the flight time of the object. A naive approach would be to increase the frame rate of the camera. However, there is an inherent trade-off between frame rate, bandwidth, and latency. Increasing the frame rate and resolution of the sensor would lead to a larger volume of data to process, and thus incur longer latencies that are detrimental to the performance of the catching system. The trade-off between high latency and high computational cost for traditional cameras is a critical obstacle on the path to achieving human-level catching performance.

Many of these problems can be avoided by using bioinspired event cameras, which are designed to emulate biological vision to enable fast perception. Their high-temporal sampling resolution, low bandwidth, high dynamic range, and asynchronous capabilities make them ideal sensors for dynamic environments. In this work, we address the question: \textit{\fernando{can we narrow the gap between robots and humans in vision-based catching tasks by using event-based vision?}}


\glspl{dvs} have been previously shown in perception systems for dodging and avoidance of dynamic obstacles~\cite{falanga2020dynamic}. We focus on the task of catching fast balls shot towards the camera, which is \fernando{a harder task as it requires both precision and speed}. Since actuators have mechanical limitations, the time allocated for perception is bounded. Under these circumstances, we have significant constraints on the latency of the perception system. Additionally, the deadline to make a control decision depends on the velocity and size of the incoming object~\cite{falanga2019how}.

\textbf{We present the first coupled event-based perception-action system capable of catching balls flying at a speed of up to  \maxspeed}. At the heart of our approach is a \textbf{novel, lightweight representation which can accurately encode event history}. The system is capable of performing inference in real-time \final{of the $x$ impact location} and issue the appropriate motion command \final{to a linear rail} to intercept the incoming ball. In summary, our contributions are:
\begin{enumerate}

    \item A new lightweight representation for events that significantly reduces the computational cost of real-time event \fernando{representation} processing. This representation outperforms both event volume and grayscale image-based perception baselines, while achieving considerably lower latency.
    \item A compact and fast event-based neural network and a robust motion estimation algorithm. The average error of the impact location is 1.9 cm.
    \item An end-to-end system to perform fast ball catching with visual perception, achieving an average success rate of \successrate  on unseen trajectories with a top speed of \maxspeed.
\end{enumerate}
\section{RELATED WORK}


\subsection{Object Catching in Robotics}
There have been multiple attempts to develop robots capable of catching a fast ball by predicting its trajectory. One of the earlier works in catching a moving object is Mousebuster~\cite{buttazzo1994mousebuster}, which intercepts an object moving at $0.7~\text{m/s}$ using a robot manipulator. More recently, visual-servoing methods~\cite{deguchi2008goal} use pixel coordinates to directly control the arm to catch an object whose flight trajectory takes approximately $1.5\,\text{s}$.
\fernando{Some works use high-speed sensors to reduce the perception latency}. Sato et al.~\cite{sato2020high} use a high-speed camera running at $500\,\text{Hz}$ to better estimate the trajectory of the object. However, achieving such a frame rate requires a high-bandwidth PCI-E interface to transfer the data. A fast color segmentation algorithm is used to estimate the position of the ball.

Another relevant area of research in object catching is sport robots. Researchers have attempted to build robots to play table tennis against real human players. The robot system in~\cite{rapp2011ping} is able to hit a vertically moving ball at $1.5\,\text{m/s}$. Monocular cameras coupled with a small baseline motion were used in \cite{7018920} to regress \fernando{the trajectory of a} ball. Recent techniques use deep learning to encode visual images of an object trajectory and predict its future \fernando{location}~\cite{7018920}. Learning-based approaches are used to predict the full trajectory of the ball given partial observations~\cite{lin2019ball}. Although the trajectory estimator in~\cite{lin2019ball} works with a ping-pong ball that travels as fast as $7\,\text{m/s}$, experiments are \fernando{performed} only in post-processing. Despite these previous attempts, the presented systems are usually too slow for real-time catching of balls returned by a real human. 

Finally, while there are videos online about robots catching balls, it is hard to assess the performance and generality of these instances.


We observe that most pieces of work in this area either 1) target low-speed motion, 2) require an external motion capture system, or 3) rely on high-bandwidth and intensive computation. In contrast, our work focuses on intercepting balls at higher speeds, using low-bandwidth event representations, capable of running in resource-constrained systems. We show a comparison between previous work and our event-based catcher in Tab.~\ref{tab:catching_methods}. Although the tasks in these papers are not exactly the same, the table captures the main challenge we address in building high-speed catching/intercepting systems.

\begin{table}[]
    \centering
    \begin{tabular}{c|c|c|c|c}
         & Real-time & \shortstack{Real \\ robot} & Monocular & Speed  \\
    \hline
     Deguchi et al.~\cite{deguchi2008goal}   & \cmark & \cmark & \xmark & $<$ 5m/s \\  
     
     Cigliano et al.~\cite{7018920} & \cmark & \cmark & \cmark & $\sim$ 4m/s \\
     
     Rapp~\cite{rapp2011ping}   & \cmark & \cmark & \xmark & $\sim$ 0.5m/s \\  
     Lin \& Huang\cite{lin2019ball}   & \xmark & \xmark & \cmark & $\sim$ 7m/s \\  
     Sato et al.~\cite{sato2020high}   & \xmark & \xmark & \cmark & $\sim$ 1.2m/s \\  
     Zhang et al.~\cite{zhang2011tracking}   & \xmark & \xmark & \cmark & $\sim$ 5 m/s \\  
     Ours   & \cmark & \cmark & \cmark & \textbf{$\sim 13$} m/s\\  
    \end{tabular}
    \vspace{.2cm}
    \caption{Feature comparison of select existing catching and table tennis robots. }
    \label{tab:catching_methods}
    \vspace{-.5cm}
\end{table}


\subsection{Event Cameras}
Event-based cameras measure asynchronous changes in \emph{log-light intensity}. 
These cameras output a set of events $E=\{e_1, e_2, ... e_n\}$, where for each event $e_i=\{x_i,y_i,p_i,t_i\}$,  $(x_i,\,y_i)$ correspond to camera pixel location, $p_i$ is the polarity (sign change of the \fernando{log-}light intensity), and $t_i$ is the time at which the light change occurs.

Event-based 2D tracking-based approaches directly estimate the motion parameters by optimizing over image gradient~\cite{gehrig2020eklt}. Tracking can be done by contrast maximization~\cite{zhu2017event,gallego2019focus} or a globally optimal search~\cite{liu2020globally}.
\fernando{\cite{monforte2020exploiting} showcases the advantages of using event cameras compared to traditional cameras to track bouncing balls using long-short-term memory (LSTM) architectures.}
Related to our task, ~\cite{glover2016event} detects ball positions by applying a Hough transform to identify full circles projected onto the image frame. These methods show promising results in tracking objects in 2D, but bringing the motion into the 3D space remains an unsolved task for event cameras. 
Learning-based approaches have been proposed to directly learn depth from monocular event data~\cite{hidalgo2020learning}. 
Attempts have also been made to directly learn the structure of the scene and the movement of the camera from the event cameras~\cite{Zhu-RSS-18}. 
Another line of research learns dense time-to-collision (TTC) from monocular event sequences~\cite{9636327}. In addition to these perception-focused works, end-to-end learning of control input from event data has enabled complicated control tasks such as UAV navigation ~\cite{9560881}.

Recently, event-based cameras have been used for high-speed dodging \cite{falanga2020dynamic, 9196877,9561290}. Early demonstrations of DVS catching have been performed for the task of goal keeping~\cite{10.3389/fnins.2013.00223}.
In this paper, we present results which further the state of the art in this line of inquiry. We take on the challenging task of catching high-speed objects, which generally requires more precision of estimating the impact location than just dodging objects.

\section{Method}

\subsection{Binary Event History Images}
\label{sec:behi}
A key challenge in dealing with event data is choosing the appropriate representation. To reduce the computational cost while preserving the necessary trajectory information of a flying object, we propose using a compact representation of events called a {\it \glsreset{behi}\gls{behi}}.
For a list of events $\{e_1, e_2,..., e_N\}$,  a \gls{behi} at time $T$ is defined as:
\claude{
\begin{align}
    I_T(x, y) &= (\sum_{i=1}^N[x_i=x, y_i=y, t_i < T]) > 0.
\end{align}
}

\begin{figure}
    \centering
    \includegraphics[width=\linewidth]{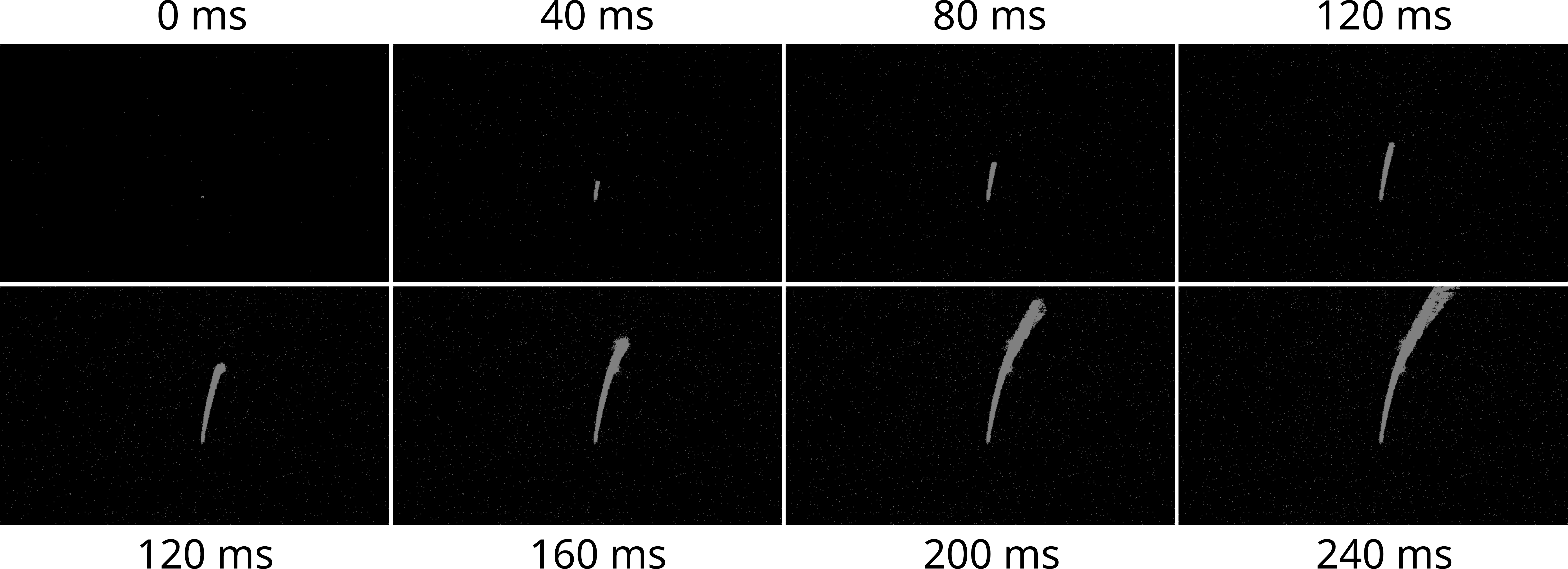}
    
    \caption{\glsreset{behi}\gls{behi} generated from a sample trajectory. From left to right, top to bottom, we show the \gls{behi} progression of a ball flying towards the camera. }
    \label{fig:binary_images}
    \vspace{-.5cm}
\end{figure}
The \gls{behi} highlights the trajectory of the ball projected onto the image plane because only pixels that have changed during the flight time are activated. In addition, such images have the same number of channels independently of the total time range of the events. Compared to grayscale images whose history requires heavy concatenation, BEHI keeps a constant-sized image that preserves the trajectory information.

BEHIs are lightweight \fernando{representations for event cameras: for sensors of resolution $W \times H$ the size required for BEHI is only ($W \times H$) bits}. On the contrary, the events volumes~\cite{Zhu-RSS-18} \fernando{require} ($C \times 2 \times H \times W \times 32$) bits, where $C$ is the number of channels for the event volume. On the other hand, if we \fernando{use} grayscale images as input, the size requirement is ($H \times W \times N \times 8$) bits, where $N$ is the number of frames used to estimate the trajectory information. 

\subsection{Learning Object Trajectory with Uncertainty}

We adopt a learning-based approach to estimate the final impact location by predicting $x$ positions and \gls{ttc} along the flight trajectory.

An important aspect of our problem consists of acquiring early event data on the incoming object, when its size in the camera is small and perception information is uncertain. To properly deal with this uncertainty, one must incorporate it into the model to produce a robust perception.

At the beginning of each trajectory, the motion is projected onto a small number of pixels in the image. This could make \fernando{initial predictions} misleading. As the object gets closer to the camera, the motion becomes more apparent and therefore can provide a more robust estimate. However, due to the limited number of data points available to the perception front-end before having to perform a catch maneuver, \fernando{we would like to use all the available data to estimate our impact position, even if} the initial data usually contain a certain amount of noise. Therefore, we adopt a confidence-driven approach to estimate the impact location by a weighted least-squares regression. 

Traditionally, such uncertainty is obtained by filtering techniques such as Kalman Filter~\cite{kalman1960new}. Inspired by~\cite{liu2020tlio}, which uses IMU-only data to estimate displacement and uncertainty, our network learns to minimize a log-likelihood function to learn the uncertainty directly.

Instead of producing a single-scalar prediction, we output the normal distribution $\mathcal{N}(d_i, \sigma_i)$ of the prediction. $d_i$ is the predicted object location in the world frame and $\sigma_i$ is the standard deviation of the distribution inversely correlated with confidence. 

\subsection{Robust Prediction Network}
\label{subsec:robust-pred-net}
In this section, we describe the prediction network to estimate the trajectory, as well as the prediction algorithm that outputs the impact location. \claude{Given a stream of events, our goal is to estimate the time and impact location of a single moving object in the scene. 

It is difficult to track the object using traditional window-based approaches for a number of reasons.
First, detecting arbitrary objects and tracking their
scales are challenging tasks, as the shape of the ball is
highly dependent on the motion itself~\cite{zhu2019motion}. Second, applying
naive tracking algorithms becomes increasingly difficult
when there is external noise.
Finally, without depth information, an accurate relative scale needs to be continuously estimated.
To overcome these challenges, we propose using
a lightweight network to learn these tasks simultaneously. }

For a given \gls{behi} from the camera, the network predicts three values: 1) \textbf{current ball x-location} in the camera frame, 2) \textbf{prediction uncertainty} and 3) \textbf{\gls{ttc}}. Previous efforts have been made to directly predict the impact location and \gls{ttc} directly from a dart trajectory~\cite{9561290}. However, such regression tasks require a significant amount of training data, since there is only a pair of such ground-truth values for each event sequence. In~\cite{9561290}, a carefully designed data augmentation method was applied by random shifting and rotation. In this work, we overcome this problem by introducing an explicit motion model to the 3D trajectory so that the network is supervised on higher frequency ground truth data. 

Due to the real-time nature of the system, our network requires to \fernando{perform at} low latency. After obtaining the \gls{behi} input, we resize it to $250\times250$
before feeding it into the network. We chose a regression network that has only 4 convolutional layers (with batch normalization) and 1 linear layer, with a total of 370,947 parameters. \claude{Each convolutional layer has 64 channels with $3\times3$ kernels. We used a stride size of 2 to decrease the resolution. Before the last linear prediction layer, average pooling is used to spatially aggregate the features. Rectified Linear Units (ReLU) are used after each BatchNorm layer as activation functions.} This simple design allows us to run inference in batches of 12 within \inferencetime. The small network demonstrates competitive performance in learning motions. We trained the network for 100 epochs with a learning rate of $2\cdot10^{-4}$ and a batch size of 16. All networks mentioned in Section~\ref{sec:results} are trained with the same architecture and hyperparameters.

\subsubsection{Object Position Regression Loss} Given the output position $d_i$ and the ground truth object location $\hat{d_i}$, the regression loss forces the network to predict the mean of the predicted normal distribution.
\begin{align}
    \mathcal{L}_{L1}^{POS}(d, \hat{d})=\frac{1}{n} \sum_{i=1}^{n}|d_{i}-\hat{d}_{i}|
\end{align}

\subsubsection{Object TTC Regression Loss} Given the output time $ttc_i$, an L1 regression loss is used to supervise the \gls{ttc} prediction from the current event time stamp, which is defined as the timestamp of the last event in the current \gls{behi}.
\begin{align}
    \mathcal{L}_{L1}^{TIME}(ttc, \hat{ttc})=\frac{1}{n} \sum_{i=1}^{n}|ttc_{i}-\hat{ttc}_{i}|
\end{align}

\subsubsection{Negative Log-Likelihood Loss} Given the output position $d_i$, standard deviation $\sigma_i$ from the network and the ground truth object $\hat{d_i}$, the negative log-likelihood loss can be computed as:
\begin{align}
\mathcal{L}_{NLL}(d, \sigma, \hat{d}) &= - \frac{1}{n} \sum^n_{i=1} \log(\frac{1}{\sigma_i \sqrt{2 \pi}} \exp(-\frac{1}{2} (\frac{d_i - \hat{d_i}}{\sigma_i})^2)) \\
&= \frac{1}{2n} \sum_{i=1}^n \log(\sigma_i) + (\frac{d_i - \hat{d_i}}{\sigma_i})^2 + cst
\end{align}

We jointly optimize the loss function by taking the weighted sum of the three functions above. The log term in $\mathcal{L}_{NLL}$ makes the network unstable if the initial mean estimate $d_i$ is inaccurate. \claude{The ball location can be learned alone with the negative log likelihood loss term. However, the convergence of the log-likelihood function depends on a good initial mean estimate and is vulnerable to outliers. Therefore, we have the L1 regression loss to help stabilize network training. $\lambda$ is $0.1$.}
\begin{align}
\mathcal{L} &= \mathcal{L}_{L1}^{POS} + \mathcal{L}_{L1}^{TIME} + \lambda \cdot \mathcal{L}_{NLL}
\end{align}



\subsection{Estimating Impact Location with Uncertainty}
\label{sec:prediction}

Given a list of noisy object position predictions ${x_i}$ and standard deviation ${{\sigma_i}^2}$, predicted \gls{ttc} ${ttc_i}$, and event frame timestamps ${t_i}$, the goal is to recover the impact location $x_{impact}$. To reduce the cost of computation needed for prediction, we assume that the trajectory model is linear with respect to time: 
\begin{align}
    \mathbf{X} = \mathbf{T} \beta + \epsilon \label{eqn:wlr}, 
\mathbf{T} = \begin{bmatrix}
1 & t_{1} \\
\multicolumn{2}{c}{\cdots} \\
1 & t_{N}
\end{bmatrix}
\end{align}
where $\epsilon$ is normally distributed and the observations are independent, we have an inverse covariance matrix. 
\begin{align}
\mathbf{\Sigma} = \left(\begin{array}{ccc}
\frac{1}{\sigma_{1}^{2}}  & \ldots & 0 \\
\vdots &  \ddots & \vdots \\
0 & \ldots & \frac{1}{\sigma_{n}^{2}}
\end{array}\right)
\end{align}
The weighted least squares~\cite{strutz2016data} based on position predictions $\mathbf{X}$ and the inverse variance matrix $\mathbf{\Sigma}$ can be written as:
\begin{align}
\hat{\beta} = \mathbf{(X^T \Sigma X) ^ {-1} X^T \Sigma T}, 
\end{align}
For the prediction of the time to collision, we take the mean of the projected impact time prediction in the future: \begin{align}
    \bar{t} = \frac{1}{N} \sum_{i=1}^N (t_i + {ttc}_i).
\end{align}
Note that $\bar{t}$ is with respect to $t_1$, which is the timestamp of the first event image. To get the impact location $x_{impact}$, we feed \gls{ttc} into the weighted regression model in Equation~\ref{eqn:wlr}:
\begin{align}
x_{impact} = 
\begin{bmatrix}
1 & \bar{t}
\end{bmatrix} \hat{\beta}.
\end{align}
\final{We use a linear model model in the x-direction under the assumption that the ball motion is ballistic, with gravity aligned with the $z$ axis. Moreover, we observe that aerodynamic effects, such as spin caused by the launcher, only affect the motion of the ball in the $z$ direction. If required, the linear model could be modified to encompass more complicated motions, such as aerodynamic effects on the $x$ axis. In Section~\ref{sec:results}, we show that simple linear models in $x$ are able to robustly estimate the impact location.}

\begin{figure}
    \centering
    \includegraphics[width=.5\textwidth]{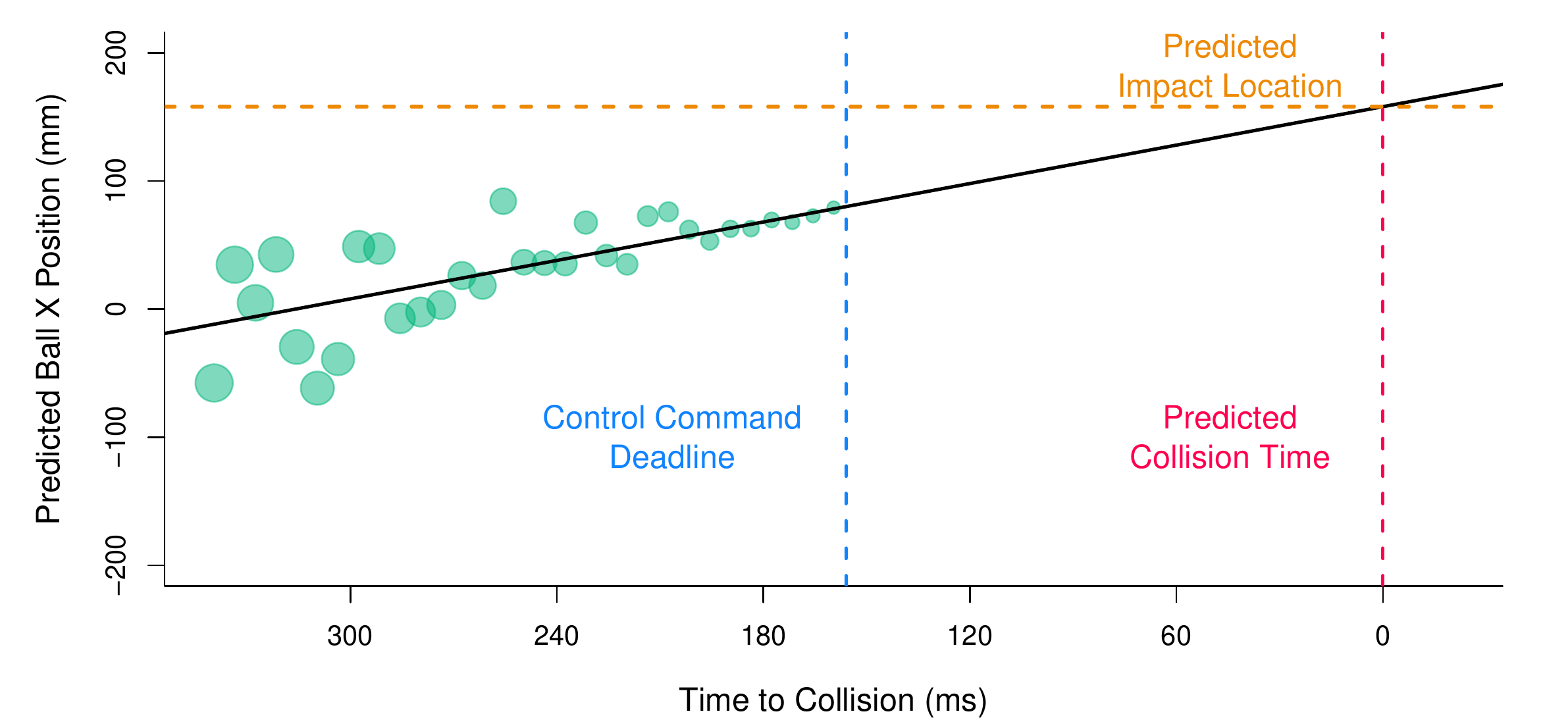}
    \caption{Robust Motion Estimator with simulated data. Green circles are the \fernando{predicted} ball locations. The size of each circle represents the amount of uncertainty of the prediction. The blue line indicates the deadline when the motor command needs to be issued. The robust motion estimation algorithm takes the these observations and predicts the final impact location.  }
    \label{fig:pred_figure}
    \vspace{-.5cm}
\end{figure}
In Figure~\ref{fig:pred_figure}, we show how this robust estimator works with simulated data. 
At the beginning, the predictions are often inaccurate with high variance. Given noisy predictions, the robust estimator produces the parameters of the linear trajectory. By using more predictions, the estimator is more robust to outliers approaching the control command deadline.



\section{Experiment Setup \label{sec:system}}

\subsection{System Design}
\label{subsec:systemdesign}
To validate the performance of the system, we performed end-to-end catching experiments. The system overview is depicted in Fig.~\ref{fig:setup} and is composed of the following elements:

\subsubsection{\textbf{Ball launcher}} it is a modified table tennis trainer (Robo-Pong 2055) where: 1) we increased the motor voltage to achieve higher ball speeds and 2) we attached \glspl{led}, a photoresistor, and an operational amplifier to generate a trigger signal when a ball is shot.  \claude{Using a trigger signal is optional, but it reduces the computational cost when using embedded hardware to  enable the operation. Alternatively, we could use a robust \gls{ttc} estimation and issue a move command with accurate timing, as explained in Section~\ref{sec:trigger}.}

\subsubsection{\textbf{Processing computer}} we use an NVIDIA Jetson NX embedded \gls{som}, with a 384-core Volta GPU. The processing computer collects events from the \gls{dvs}, generates the BEHIs, runs the network inference, and sends the position command index to the motor controller using its GPIO. We use the Jetson for two reasons: 1) the fast GPIO is crucial in our perception-action loop. The GPIO input / output latency of this platform is sub $1\,\text{ms}$. 2) The platform requires only $15\,\text{W}$, which is important in the context of low-power perception with event cameras. \fernando{The only disadvantage of using the Jetson NX is that this platform only allows for a \emph{single-shot} inference, and thus it requires a trigger signal.}  
\subsubsection{\textbf{Controller and actuator}} We use an off-the-shelf Festo CMMP-AS motor controller, with an EMME-AS motor and a EGC toothed belt axis. \fernando{The actuator is located perpendicular to the motion of the ball.} It has a motion span of $600\,\text{mm}$ ($300\,\text{mm}$ from the center position), and it is capable of achieving a velocity of $10\,\text{m/s}$ and an acceleration of  $50\,\text{m/s}^2$. The target motions are precomputed in a look-up table at intervals of $100\,\text{m/s}$ using a constant-jerk trajectory planner. We attached a rim with a net that has a height of $400\,\text{mm}$ and a width of $240\,\text{mm}$ (including the rim).

The hard latency constraint of our mechanical system \fernando{(see Figure~\ref{fig:pred_figure})} comes from the actuator. The constant-jerk trajectory planner is only capable of $\pm 1300\,{\text{m/s}^3}$. As we would like to reach distances as far as $400\,\text{mm}$ from the center of the linear actuator, including the rim. With our controller, it takes $160\,\text{ms}$ to move $283\,\text{mm}$. As the whole motion of the ball takes at least $310\,\text{ms}$ (see Sec.~\ref{sec:datacollection}), this leaves us with $150\,\text{ms}$ to execute our perception and action algorithms.

\begin{table*}[t]
    \centering
    \begin{tabular}{c|c|c|c|c}
            &  Ball Location (mm) & TTC (ms) & Impact Location (mm) & Collision Time (ms) \\
            \hline
        BEHI            & \textbf{7.809$\pm$4.208}   & \textbf{8.990$\pm$7.310} & \textbf{19.000$\pm$14.878} & \textbf{7.950$\pm$6.920} \\
        Event Volume    & 18.340$\pm$5.523     & 39.380$\pm$12.453 & 21.600$\pm$14.34 & 39.380$\pm$10.654
\\
        Grayscale Image & 32.639$\pm$17.311    & 57.047$\pm$22.242 & 94.099$\pm$49.154 & 56.792$\pm$14.849 \\
        \hline
        \claude{BEHI (BG)} & \claude{20.554$\pm$10.010 }& \claude{14.437$\pm$12.709} & \claude{59.586$\pm$53.165} & \claude{12.292$\pm$11.371} \\
        \claude{Event Volume (BG)} & \claude{19.296$\pm$8.441} & \claude{36.214$\pm$12.478} & \claude{74.754$\pm$49.308} & \claude{36.214$\pm$10.102} \\
    \end{tabular}
    \vspace{.2cm}
    \caption{Average error of location, TTC, impact location and collision time on 20 held-out unseen trajectories. \claude{The error shown corresponds to one standard deviation. ``BG" indicates that there is background motion in the scene.}}
    \label{tab:performance}
    \vspace{-.5cm}
\end{table*}

\begin{figure}
    \centering
    \includegraphics[width=.78\linewidth]{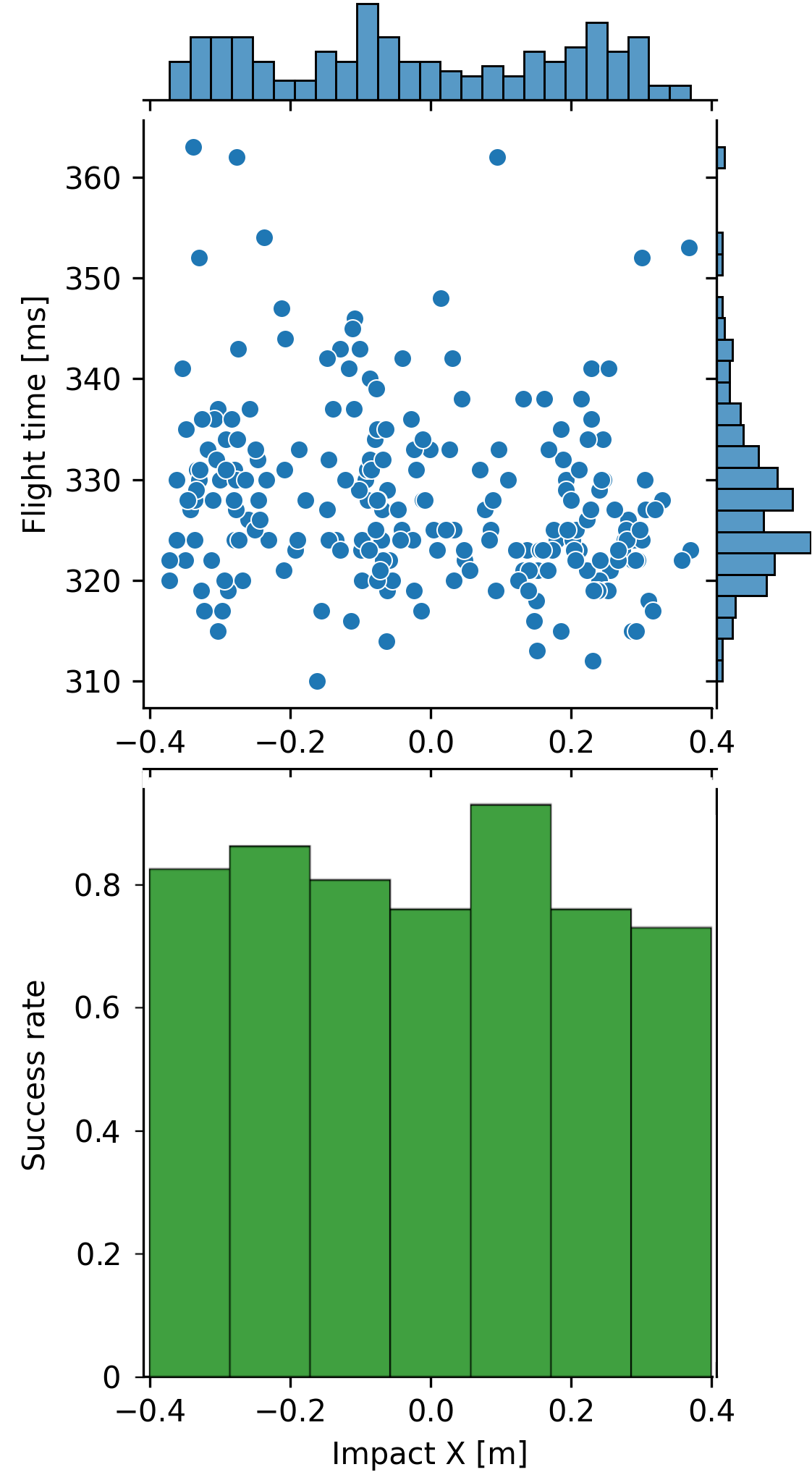}
    \caption{\underline{Top}: Impact location and flight time for the different data points in our collected data set, with their respective histograms. We aim to cover different shooting angles and speed, allowing the network to learn unbiased estimation of ball motions. 
    \underline{Bottom:} 
    experimental end-to-end success rates for different impact locations of the ball.}
    \label{fig:impact_loc}
    \vspace{-.5cm}
\end{figure}

\subsubsection{\textbf{Perception-Action}}
The main sensor of our perception system is an Inivation DVXplorer \gls{dvs}, featuring a Samsung \gls{dvs} Gen3 sensor~\fernando{\cite{ryu2019industrial}}. \fernando{The \gls{dvs} is located in front of the actuator, pointing toward the ball launcher}.
Our DVS driver generates a \gls{behi} from the events as soon as we receive them\fernando{, every $10\,\text{ms}$}. Then
on the inference program, we assemble our batch of \fernando{12} \glspl{behi} and make a prediction. Next, we perform regression to estimate the impact position and decide which is the best position command for the linear actuator to intercept the ball. Finally, the target impact location is then sent to the controller to move the linear actuator.


\subsection{Data Collection and Preprocessing}
\label{sec:datacollection}

\fernando{While there exist simulators for event cameras, we have observed that they may not produce accurate representations of the sensor. For this reason, we decided to train our network using only real-world data}.
The data collection process was carried out using a motion capture system. We wrapped our balls in reflective tape and recorded ball positions, events, and global shutter grayscale images for comparison (using a Chameleon3 camera working at 76 FPS). We manually synchronized the different sensors in the data set. A total of 235 data points were collected, and we show the distribution of flight time and impact location in Fig.~\ref{fig:impact_loc}.

As the ball motion is fast, motion capture systems may lose track intermittently for a few milliseconds. To solve this problem, we performed a second-order polynomial fitting of the ball position. This has the added benefit of simplifying the ball motion pattern, making it easier for the network to learn. Furthermore, interpolation provides intermediate position data points that can be used during the training process, as described in Sec.~\ref{sec:prediction}.

\claude{
\subsection{Triggering the Motion with Network Prediction}
\label{sec:trigger}

The latency of our system is bounded by the mechanical limitations of our actuators (Section~\ref{subsec:systemdesign}).  Moreover, once we send the motion command, it is not possible to make any corrections afterward. Therefore, it is critical to issue the motion command to the rail on time. For this, we have two options: 1) we can use a trigger signal to start processing the events as soon as the ball is shot, or 2) we can run our network continuously and wait for the last predicted TTC to fall within a certain amount of time reserved for moving the rail. \final{In Tab.~\ref{tab:trigger}}, we show a comparison of the performance of using these two methods. To ensure enough time for the rail motion, we continuously run the network until the last predicted TTC is within $210\,\text{ms}$. We then stop the network inference and issue a command based on the predictions up to this point.
 
Despite the similar performance of using trigger signals by thresholding the predicted TTC, this method relies on running our inference pipeline every time we receive a \gls{behi}. \final{The experiments in Tab.~\ref{tab:trigger} are performed on a much more powerful laptop computer with a mobile RTX 3080 GPU to simulate pipelining.}  Our computing platform is a lightweight Jetson NX, which does not have such computational power. Therefore, we relax this computational requirement by using a trigger signal.
}

\section{Experimental Results}

\begin{figure*}
    \centering
    \includegraphics[width=\linewidth]{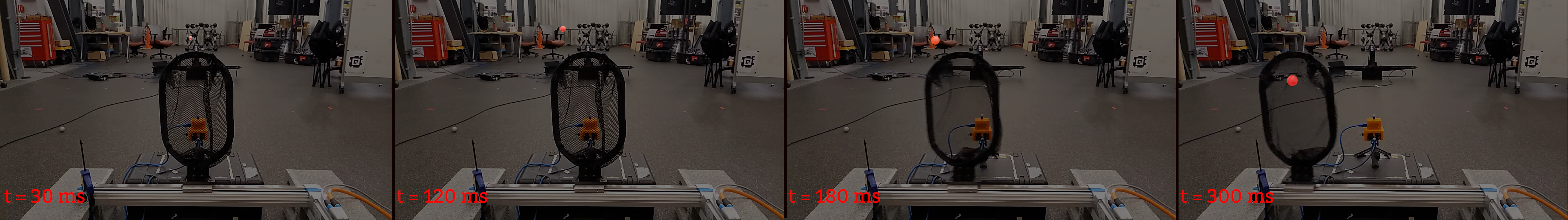}
    \caption{Catching sequence for an example ball launch. We collect events every $10\,\text{ms}$, for a total of  $120\,\text{ms}$. During this time, the rail does not perform any motion. Next, we perform inference on the embedded computer. Finally, we select a target position to perform a one-shot motion to catch the ball. }
    \label{fig:endtoend}
\end{figure*}

\label{sec:results}
In this section, we present the quantitative results of the front-end perception of our system, as well as the end-to-end object catching results.

\subsection{Perception}


To provide baselines for the perception module, we evaluated the performance on two other input modalities: event volumes and grayscale images. For event volumes, we follow the formulation in~\cite{Zhu-RSS-18} to generate event volumes with 10 temporal bins for each polarity. For a set of events $\{e_i = (x_i, y_i, t_i, p_i)\}$, an event volume $V_p(x, y, t)$ is defined as
\begin{align}
t_{i}^{*} &=(B-1)\left(t_{i}-t_{0}\right) /\left(t_{N}-t_{1}\right) \\
V_p(x, y, t) &=\sum_{i} p_{i} k_{b}\left(x-x_{i}\right) k_{b}\left(y-y_{i}\right) k_{b}\left(t-t_{i}^{*}\right) \\
k_{b}(a) &=\max (0,1-|a|)
\end{align}
where $k_b$ is a bilinear interpolation kernel. This representation has been widely used in flow estimation because it preserves the temporal gradient of events.

For grayscale images, due to the framed nature of the camera, we cannot directly encode variable-sized frames in a Convolutional Neural Network without computationally heavy recurrent structures. 
In addition, event volumes and stacked image frames both have multiple input channels with non-binary inputs (unsigned integer or floating point). These properties make it challenging to run inference in real time, even with GPU acceleration. 
Limited by the longer network inference time and data transmission latency of using grayscale images, we have time to collect only 8 images for prediction.

\begin{table}[t]
    \centering
    \claude{
    \begin{tabular}{c|c|c}
            &  Impact Location (mm) & Collision Time (ms) \\
            \hline
        Hardware Trigger    & \textbf{19.000$\pm$14.878} & \textbf{7.950$\pm$6.920} \\
        Network TTC Trigger & 22.755$\pm$21.728 & 8.216$\pm$6.757 \\
    \end{tabular}
    \vspace{.2cm}
    \caption{Influence of the trigger signal in the perception performance of our system. }
    \label{tab:trigger} 
    }
    \vspace{-.5cm}
\end{table}

\begin{figure}[t]
    \centering
    \includegraphics[width=.85\linewidth]{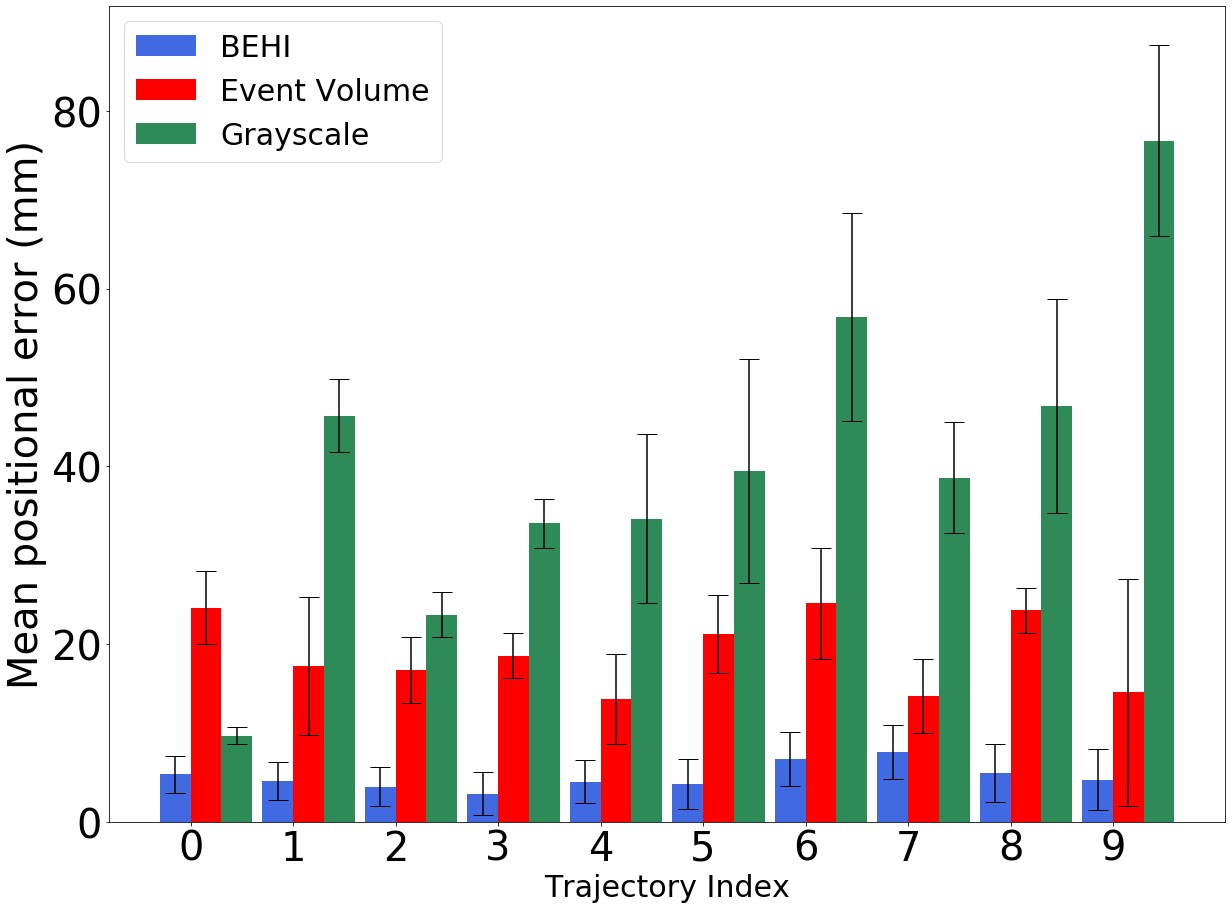}
    \caption{Average error of the object position prediction (reported in meters) on a subset of the 20 unseen test trajectories.}
    \label{fig:pos_figure}
    \vspace{-5mm}
\end{figure}

We evaluate the perception component of the system by evaluating two errors: 1) per frame error and 2) per trajectory error. 
We provide four error metrics for the following predictions.
\begin{enumerate}
    \item Ball location (per frame): the predicted location of the object in the camera frame. This is evaluated in all data before the 160~ms deadline.
    \item \gls{ttc} (per frame): the predicted time to collision of the object. This is also evaluated in all data before the $160\,\text{ms}$ decision deadline.
    \item Impact location (per trajectory): the predicted impact location (x coordinate) when the object hits the image plane. This prediction combines time-to-collision and ball location estimates.
    \item Collision time (per trajectory): the time from the ``trigger time" to the time when the ball hits the image plane.
\end{enumerate}
We provide the quantitative evaluation in Tab.~\ref{tab:performance}. 
Additionally, in Fig.~\ref{fig:pos_figure} we report the ball location error per frame among the three input representations for a subset of unseen test trajectories.

 We observe that the network trained with \gls{behi} consistently outperforms the other two baseline networks on different unseen trajectories. Grayscale images contain not only the motion trails, but also the background pixels. With the limited amount of training data, the neural network needs to learn to identify the motion in the cluttered background. In contrast, as shown in Fig.~\ref{fig:binary_images}, the trajectory of the ball is apparent in the BEHI, which could reduce the network capacity requirement of the task. \gls{behi} does better on average than the heavier event volume representation despite its compact size. \final{We observe that more informative representations, such as event volumes, require more network capacity for learning the motion. The fundamental trade-off between network size and performance leads us to use a compressed representation such as BEHI} \claude{In this task, the impact location error for gray scale cameras has a mean of 9\final{.4} cm with a standard deviation of 5 cm. This means that many of the predictions may have a positional error greater than 14 cm.  As our fast rail uses a discretized positional controller with precomputed trajectories, a slight shift in the predicted position could cause the robot to move to a wrong position, further magnifying the inaccuracy.} 

\claude{An important component of the system is the predicted uncertainty, which is learned with an unsupervised loss. We conducted an ablation study to analyze the effect of uncertainty. The reader should note that this only changes the impact location prediction results. In our experiments, the impact location for BEHI becomes $30.078\pm31.378\,\text{mm}$ when uncertainty is ignored, as opposed to $19.000\pm14.878\,\text{mm}$ when weighted with uncertainty.}
\vspace{-0.2cm}

\begin{figure}[t]
    \centering
    \includegraphics[width=1.\linewidth]{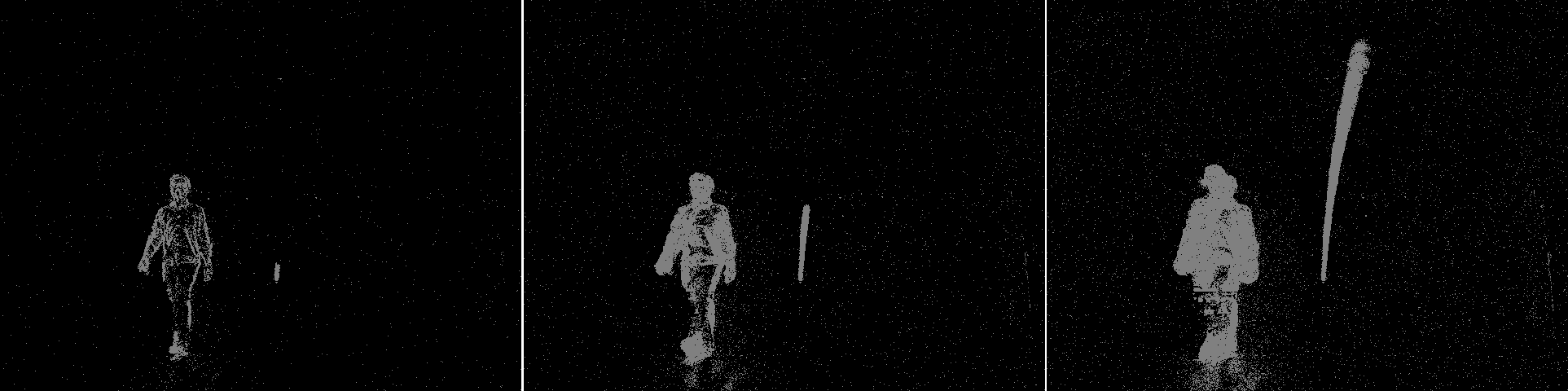}
    \caption{Example BEHIs with background motion.}
    \label{fig:bg_fig}
    \vspace{-.5cm}
\end{figure}

\fernando{
\subsection{Influence of Background Motion}
\label{sec:discuss}
In this section, we present additional experiments to test the performance of the network under significant background motion. As performing data collection of flying balls with background motion is challenging, we propose the following evaluation scheme for the perception pipeline: we first record events of people motion at the same location where we performed the experiments. Later, we merged flying ball events with the background motion events. 

We trained our networks in this new dataset and measured performance in unseen trajectories. Example BEHI images for the augmented sequences are shown in Fig.~\ref{fig:bg_fig}. We report this performance in Tab.~\ref{tab:performance}. In these extremely challenging cases, traditional blob detection and tracking algorithms are inadequate due to dominating noise. 

Although some performance degradation is observed due to background movements, the system is able to predict with a positional error of 6 cm. The \gls{ttc} and the collision time are mostly unaffected by this change. This increase is expected because background events make it more difficult to segment the ball motion. 
However, the latency of the data acquisition and inference pipelines is not affected. From Tab.~ \ref{tab:performance}, we can see that \gls{behi} underperforms event volumes in this scenario. The additional time channels of the event volumes compared to \gls{behi} could allow the network to group actions more effectively.

%
}


\subsection{End-to-end Performance}
To assess the performance of our system, we performed 120 shots targeting different locations of the end actuator, uniformly covering the whole range of motion of the linear actuator, and we tried to catch them.  We show an example of catching sequence in Fig.~\ref{fig:endtoend}.
After each motion, we return the linear actuator to the center position. The results are summarized in Fig.~\ref{fig:impact_loc}.

We observe that the average success rate is $81\,\%$, and the lowest success rate is $73\,\%$. For reference, we could expect that an average success rate for random motions will be $28.5\%$. We should note that the failure cases are different depending on the impact location: for extreme impact locations and fast balls, the latency of our perception algorithm is sometimes higher than our deadline. In rare cases, our perception system sometimes estimates the wrong position and misses the ball.

\section{Discussion}

Overall, our system is designed to close the gap between robots and humans in the task of object catching. Acosta et al.~\cite{acosta2003ping} highlighted the challenges in this task for robots using traditional cameras.
Although the high sampling rate and low latency of event cameras help solve this problem, a perception pipeline must be carefully designed to minimize the latency of each system component. Among these designs, we describe in detail the \gls{behi} representation and a lightweight event network.

From the quantitative evaluation in Tab.~\ref{tab:performance}, we observe a significant increase in perception performance using the proposed BEHI representation, compared with images and event volumes. 
\fernando{We have also showed that the \gls{behi} is robust to simple background motion.} 
For robust prediction of the trajectory and time to collision, we present an uncertainty-driven approach for fusing the high-frequency predictions from the network. This approach is necessary to compensate for inaccurate predictions from the lightweight network. The uncertainty of each prediction is learned unsupervised from the distribution of the training data and therefore does not require additional labeling. 
Using this approach, the perception algorithm is able to maintain an increasingly robust estimate of the impact location at a low latency. \fernando{Our system does not require to pre-calibrate the sensors, as the calibration is learned from the training data}. \final{Since we do not explicitly model the calibration and configuration of objects into the method, knowledge such as ball shapes and lens parameters that are essential for motion estimation may not easily transfer to new scenes. Therefore, retraining the network is expected for a new experimental setup.}

We should note that due to the mechanical limitations of our position controller, we could only issue single-shot commands to the rail. One could imagine a faster controller is able to ``follow" the prediction from the network over time, which would allow the actuator more time for movement.
\fernando{Moreover, we have assumed a prior in the motion of the flying object (linear) during our experiments, as explained in \ref{subsec:robust-pred-net}.}

\section{Conclusion}
This work investigated the problem of catching high-speed balls using event-based sensors. Through our study, we were able to show that event-based cameras are an attractive sensor for this task compared to frame-based sensors. Additionally, we demonstrated a full-scale system of perception, planning, and action to achieve catching at a top speed of \maxspeed. Both of these achievements present interesting routes for further high-speed catching systems. 

One direction of future investigation is solving the vision task of perceiving the terminal object state given a different camera viewpoint. This approach would enable the development of mobile robots that can perform the object catching task. \fernando{Moreover, we have not analyzed in this paper the effect of sensor egomotion. Although simple rotations can be easily compensated using an on-board IMU \cite{falanga2020dynamic}, general motion segmentation using monocular event cameras remains challenging with the tight time constraints presented in this paper.} Another future direction is to reduce system latency through the development of specialized hardware for edge inference.  This would permit the deployment of our method on a resource-constrained system running on minimal energy. All in all, we seek to create the next generation of low-latency robot systems that can respond and react to the dynamic environment around them.

\final{
\section{Acknowledgement}
We gratefully acknowledge Samsung AI 2021-2022 Award to the University of Pennsylvania.
}


\bibliographystyle{IEEEtran}
\bibliography{IEEEabrv, bib}

\end{document}